\documentclass[conference]{IEEEtran}
\IEEEoverridecommandlockouts
\usepackage{cite}
\usepackage{amsmath,amssymb}
\usepackage{algorithmic}
\usepackage{graphicx}
\usepackage{textcomp}
\usepackage{xcolor}
\usepackage{booktabs}
\usepackage{multirow}
\usepackage{comment}
\usepackage{subcaption}
\usepackage{fancyhdr}

\begin{document}
\title{A Self-Supervised Approach for Enhanced Feature Representations in Object Detection Tasks
\thanks{Corresponding author: Beatriz Remeseiro (bremeseiro@uniovi.es).\\
Grant PID2019-109238GB-C21 funded by MICIU/AEI/10.13039/ 501100011033.\\
© 2024 IEEE. Personal use of this material is permitted. Permission from IEEE must be obtained for all other uses, in any current or future media, including reprinting/republishing this material for advertising or promotional purposes, creating new collective works, for resale or redistribution to servers or lists, or reuse of any copyrighted component of this work in other works.}
}

\author{\IEEEauthorblockN{Santiago C. Vilabella}
\IEEEauthorblockA{\textit{Menéndez Pelayo International University} \\
Spain \\
santiagocastrovilabella@gmail.com}
\and
\IEEEauthorblockN{Pablo Pérez-Núñez}
\IEEEauthorblockA{\textit{Artificial Intelligence Center} \\
\textit{University of Oviedo}\\
Gijón, Spain \\
pabloperez@uniovi.es}
\and
\IEEEauthorblockN{Beatriz Remeseiro}
\IEEEauthorblockA{\textit{Artificial Intelligence Center} \\
\textit{University of Oviedo}\\
Gijón, Spain \\
bremeseiro@uniovi.es}
}

\maketitle

\begin{abstract}
In the fast-evolving field of artificial intelligence, where models are increasingly growing in complexity and size, the availability of labeled data for training deep learning models has become a significant challenge. Addressing complex problems like object detection demands considerable time and resources for data labeling to achieve meaningful results. For companies developing such applications, this entails extensive investment in highly skilled personnel or costly outsourcing. This research work aims to demonstrate that enhancing feature extractors can substantially alleviate this challenge, enabling models to learn more effective representations with less labeled data.  Utilizing a self-supervised learning strategy, we present a model trained on unlabeled data that outperforms state-of-the-art feature extractors pre-trained on ImageNet and particularly designed for object detection tasks. Moreover, the results demonstrate that our approach encourages the model to focus on the most relevant aspects of an object, thus achieving better feature representations and, therefore, reinforcing its reliability and robustness. 
\end{abstract}

\begin{IEEEkeywords}
self-supervised learning, feature extraction, object detection, deep learning, representation learning
\end{IEEEkeywords}

\section{Introduction} \label{sec:introduction}

In recent years, artificial intelligence has seen unbelievable advances, going from theoretical discoveries to practical solutions that give answers to real-world problems. Within this huge landscape, deep learning has become the spearhead of innovation, opening the door to advancements in almost every industry. Computer vision, one of the disciplines where deep learning has had more impact, has seen continuous improvements over the last years, taking it further each step and achieving incredible results. This has been made possible by advances in high-quality data and more sophisticated architectures, such as convolutional neural networks (CNNs)~\cite{o2015introduction, russell2010artificial} or, more recently, transformers~\cite{vaswani2017attention}. However, the application of these technologies in real-world scenarios also presents its own set of practical challenges. 

As deep learning models continue to grow in complexity and scale, the acquisition of high-quality data has quickly become a critical bottleneck. Object detection, an advanced computer vision task, aims to localize and classify objects within an image. Unlike simpler tasks that require only a class label per image, object detection necessitates both a class label and bounding box coordinates for each object present in the image. This level of granularity makes data labeling a labor-intensive and expensive procedure. For businesses, this typically involves allocating extensive hours of work to skilled employees or outsourcing the task to external companies at considerable expenses. Consequently, there is an urgent need to explore alternative approaches that could reduce the reliance of object detection applications on extensively labeled datasets. 

Regarding network architectures, CNNs can be divided into two components: the backbone and the output module. The backbone, also known as the feature extractor, is responsible for learning to detect and extract relevant features from the input data. The knowledge acquired during the learning process is subsequently used by the output module to adapt to a specific downstream task, such as classification or object detection. Scientific research in this field has demonstrated that by improving feature extractors, models can learn more insightful representations, thereby reducing the need for extensive labeled samples for effective training~\cite{o2015introduction, vaswani2017attention}. An important strategy for enhancing efficiency in this area is called transfer learning~\cite{russell2010artificial, goodfellow2016deep}. This approach involves using the information extracted from a backbone pre-trained on a large labeled dataset, such as ImageNet~\cite{russakovsky2015imagenet}, to improve the performance of a different model on a more specific task that requires a smaller dataset. However, in the context of computer vision applications, these backbones are often pre-trained for classification tasks and, therefore, tend to capture a limited set of distinctive features from the most relevant object in the image~\cite{bu2021gaia}. This often leads to poorer representations, thus reducing the efficiency of the model for localization tasks.

Recent advances in self-supervised learning (SSL) techniques have revealed that models with a robust backbone pre-trained on large amounts of unlabeled data and then fine-tuned on smaller labeled datasets can achieve performance that matches or exceeds that of models trained solely on labeled data. The objective of this research work is to explore the potential of self-supervised learning in the development of a robust feature extractor for object detection, with a focus on reducing both the complexity of the model and the amount of labeled data required to train downstream tasks. Using two well-known datasets, we demonstrated that the backbone trained using the proposed SSL strategy outperforms existing state-of-the-art methods in object localization tasks. Moreover, we assessed the qualitative aspects of representation learning, showing that our approach enables the model to focus on more relevant parts of the objects, obtaining richer representations that capture the objects with greater detail.

The main contributions of this work are summarized below:
\begin{itemize}
    \item \textbf{Enhanced feature extractors.} We present an enhanced feature extractor that can substantially increase the object localization performance.
    \item \textbf{Self-supervised learning.} The proposed feature extraction model is trained without requiring any labeled data.
    \item \textbf{Relevant aspects.} Unlike the state-of-the-art models, the resulting representations focus on the most relevant objects within an image.
\end{itemize}

The remainder of this manuscript is organized as follows. Section~\ref{sec:related_work} introduces the background of object detection and provides an overview of the literature in the field of self-supervised learning. Section~\ref{sec:methodology} showcases the proposed methodology. Section~\ref{sec:experimental_setup} describes the experimental setup. Section~\ref{sec:results} analyzes the results obtained. Finally, Section~\ref{sec:conclusions} shows the conclusions and the future lines of research. 

\section{Related work} \label{sec:related_work}

This section presents a brief background to the research, describing the problem of object detection and how deep learning has transformed it. This is followed by the limitations of traditional methods and the emergence of self-supervised learning as a viable solution.




\subsection{Object detection}

Within computer vision, object detection, the ability to locate and recognize objects within an image, has always been one of the more challenging applications. Before the usage of deep learning in the field, traditional methods like Haar cascades~\cite{soo2014object} and histogram of oriented gradients~\cite{dalal2005histograms} were very popular. However, they did not generalize well across many object classes or real-world environmental conditions. 

These limitations were overcome by the appearance of deep learning methods~\cite{zhao2019object}, which came with the disadvantages of requiring large amounts of labeled data and computational resources. The most common way of training deep learning models is by using a supervised learning approach. This requires large amounts of images that need to be acquired and labeled, maintaining quality and consistency to represent the variability of the problem.

An important aspect of deep learning methods lies in their capability to learn and generate high-quality representations from input data. This entails extracting relevant features from an image without human intervention. To obtain this feature extractor, it is required to have a dataset that represents in the most reliable way possible all the scenarios that the application will face and to be able to generalize to new unseen examples. This requirement can be a major limitation due to the difficulty of acquiring and manually labeling such amounts of data, especially for object detection tasks where the annotations are not only the labels but also the bounding boxes~\cite{adhikari2021iterative}.


\subsection{Self-supervised learning}

Scientists have recently introduced several unsupervised learning techniques aimed at mitigating the bottleneck associated with the acquisition and annotation of high-quality data. This is the case of self-supervised learning~\cite{balestriero2023cookbook}, which aims to learn meaningful representations from large amounts of unlabeled data. The main idea behind SSL is to use unlabeled data to train a robust feature extractor capable of representing data excellently and generalizing effectively across various downstream tasks. This seemingly straightforward idea is very powerful, especially given our capacity to generate large datasets that could be used without relying on human labeling. 

A common strategy to train such a feature extractor involves defining a model with a pretext task, typically distinct from the final downstream task. This pretext task is exclusively employed during the pre-training phase and its objective is to capture some hidden properties of the data. In natural language processing tasks, this can be some missing word from a sentence or to organize several unordered words. In computer vision, predicting missing parts of images is impractical due to the high dimensionality of data. Researchers have demonstrated that by applying transformations to the original images, models can effectively learn to recognize features and patterns just by using information derived from those modifications. Simple methods suggest applying a transformation to each image and making the model predict certain parameters associated with that transformation. For instance, Gidaris et al.~\cite{gidaris2018unsupervised} proposed a method in which each image is rotated, and the pretext task involves predicting the rotation angle. This approach enables the model to acquire insights into the image under various conditions, such as rotation. However, the features learned by the model also evolve with the applied modification, making these features covariant to the image transformations. While such methods exhibit powerful capabilities in certain downstream tasks, they may fall short in more intricate ones, like object detection. In tasks where the model must recognize an object regardless of its position or size, we require a technique that extracts invariant features concerning image transformations~\cite{misra2020self}. This implies that, even when the image is rotated or transformed in any other way, the relevant features learned by the model remain consistent.

Contrastive learning techniques~\cite{chen2020simple} also aim to learn feature representations by incorporating transformations. In this case, the original image serves as the reference, while the transformed versions act as the positive pair. The model learns to create similar representations for both transformed images and thus, recognizes the important features of the objects, independently of the transformations performed on them. For this approach, various types of transformations can be employed. For example, geometric transformations help the model to recognize representations that remain invariant to changes in position, size, or shape of the object; color transformations make the model more robust to color alterations; and other transformations like blurring or distortion help with variability in lighting conditions. The underlying idea behind these methods is to learn representations of images that remain invariant across a wide range of conditions.

Chen et al.~\cite{chen2020simple} introduced SimCLR, a framework for simultaneous contrastive learning of visual representations. This approach consists in training the model for a pretext task, using a loss function that minimizes the distance between the feature vectors of the positive pairs while simultaneously maximizing it for the negative pairs (i.e., the modified versions of other images within the same batch). Throughout the training process, the transformed versions of each image are pulled closer to each other while being pushed apart from the transformed versions of every other image in the batch. 

Most of the contributions in the field of self-supervised learning primarily focus on evaluating the performance of these methods in image classification tasks, with only a scant focus on their impact in object detection~\cite{balestriero2023cookbook}. While these contributions are predominantly geared towards complex applications, the objective of our research is to emphasize not only the potential of these methods in classification problems but also their applicability in localization tasks. We aim to gain a deeper understanding of how effectively these feature extractors acquire meaningful object representations and compare them to the representations obtained through the models typically employed in transfer learning, which are pre-trained on costly supervised datasets. 

\section{Methodology} \label{sec:methodology}

This section presents the proposed methodology to enhance feature extraction in object detection using SSL techniques. First, we describe the SSL feature extractor and the pre-training procedure. Next, we close the section describing the object detector that uses the feature extractor as the backbone.

\subsection{SSL feature extraction}

The feature extractor was trained using SimCLR~\cite{chen2020simple} as the primary algorithm. The main reason for selecting this SSL technique is its efficiency in acquiring complex representations, as compared to simpler SSL techniques that employ simplistic augmentations like image rotation. SimCLR offers a method that excels in handling different image patterns, such as perspectives, object sizes, or light conditions, which may be usual in real-world applications.


The effectiveness of SimCLR relies on the concept of the contrastive loss function. This function is designed to bring positive pairs, which are two augmented versions of the same image, closer together in the feature space (making their representations more similar), while simultaneously pushing apart negative pairs, which are two augmented versions of different images (making them different). By using contrastive loss for training, models learn to recognize the augmented versions from the same image as similar, thereby extracting robust and generalized features. 

\begin{figure}[!htbp]
  \includegraphics[width=\linewidth]{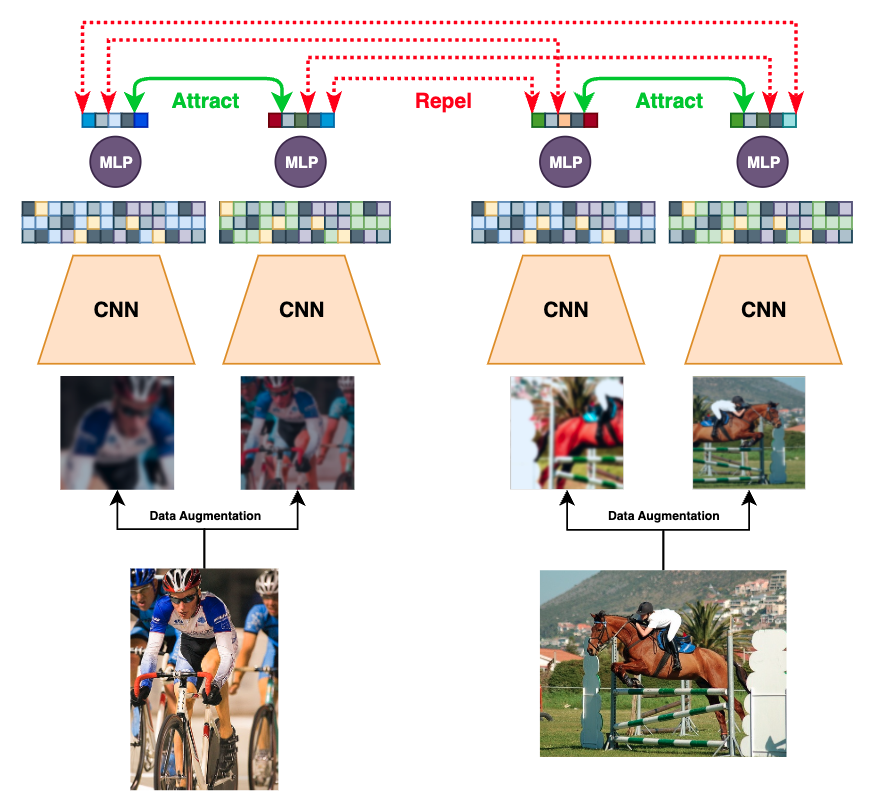}
  \caption{Simplification of the SimCLR algorithm. For the sake of clarity, we have represented the model multiple times, each applied to a different input image.}
  \label{fig:simclr}
\end{figure}

Fig.~\ref{fig:simclr} shows a simplified explanation of how the algorithm works. On the left, we can see an image of a \textit{cyclist}, which passes through a data augmentation layer that creates two new images by modifying the original one. The network is then asked to take these augmented images and generate two feature vectors.
To maximize the similarity between the two representations of the same input image, while minimizing similarities with all other representations, we use an implementation of the InfoNCE loss function~\cite{oord2018representation}.
This function, defined as
\begin{equation}\label{eq:loss_fn}
\ell_{i,j} = -\log \frac{\exp(\text{sim}(z_i, z_j) / \tau)}{\sum_{k=1}^{N} \exp(\text{sim}(z_i, z_k) / \tau)},
\end{equation} 
\noindent compares the similarity of the two augmented versions of the image, $z_i, z_j$, to the similarity of one of them to any other representation in the batch, $z_j$. In this equation, $\text{sim}$ represents the cosine similarity function, $\tau$ is the parameter in charge of adjusting the strictness of the model, and $N$ is the total number of samples in the batch.


Regarding the architecture used to implement the SimCLR algorithm, we used the convolutional base of the EfficientNet B1 model~\cite{tan2019efficientnet} as the backbone feature extractor. This base network is responsible for extracting the feature vectors from the images generated through data augmentation. Then, the output module of the original EfficientNet architecture was turned into a multi-layer perceptron (MLP), responsible for mapping the representations into a space where the InfoNCE loss is applied, maximizing or minimizing their similarities. Following the original SimCLR implementation, this projection head is a two-layer MLP.


\subsection{Object detection}

Once the feature extractor is trained following the proposed SSL strategy, it can be used to obtain the final object detector. 
The idea here lies in transferring the knowledge from the pre-trained feature extractor to the downstream task, thus requiring less data during the learning process and a simpler model. 

The architecture of the object detector has been intentionally simplified to emphasize the quality of the feature extractor trained with SimCLR. A common approach to assessing whether the model can generalize effectively to new images involves the use of a single linear fully connected (FC) layer after the backbone. This layer maps the representations from the feature extractor to the predictions. To ensure that the pre-trained weights remain unchanged during this phase, 
the weights of the backbone are frozen, and only the final layer is trained. This constraint implies that the model will perform well only if the representations generated by the pre-trained feature extractor describe all the relevant features of the image. 


As this research is focused on object detection, we require the model to handle both classification and localization tasks. For this reason, we added two downstream heads to the model architecture. The classification head maps the output features of the last convolutional layer on the feature extractor to a linear FC layer, with the number of units corresponding to the classes in the dataset. Regarding the localization head, it maps the output features to a linear FC layer with four units, each responsible for extracting one coordinate of the bounding box. Finally, it is worth noting that the simplicity of the architecture poses a significant challenge in solving the problem effectively if the feature extractor has not learned to generate sufficiently accurate representations of the images. 

The loss function of the object detector must consider both the classification and localization outputs. For the classification task, we selected the categorical cross-entropy loss~\cite{mao2023cross}, which provides a probabilistic interpretation of the predictions. This function, which has proven to be useful for computer vision tasks with several classes, can be defined as

\begin{equation}
\text{CCE} = -\sum_{i} y_i \cdot \log(p_i),
\end{equation}

\noindent where \(y_i\) stands for the ground truth value of the sample \(i\) and $p_i$ is the probability, as predicted by the model, that the sample belongs to class $i$.

For the localization task, we used the distance-IoU loss~\cite{zheng2020distance} that compares the bounding box generated by the model with the one from the ground truth and includes a penalty term that directly minimizes the distance between the central points.
This loss function is defined as

\begin{equation}
\text{DIoU} = 1 - \frac{|B \cap B_{gt}|}{|B \cup  B_{gt}|} + \frac{\rho ^2 (\mathbf{b}, \mathbf{b}^{gt})}{c^2},
\end{equation}

\noindent where $B$ and $B_{gt}$ represent the predicted and the ground-truth bounding boxes, $\mathbf{b}$ and $\mathbf{b}^{gt}$ denote their central points, $\rho(\dot)$ is the Euclidean distance, and $c$ is the diagonal length of the smallest enclosing box covering the two bounding boxes.

Finally, these two loss functions are combined into a single one to be used by the object detector
\begin{equation} \label{eq:object_loss}
\mathcal{L} = \alpha * \text{CCE} + (1-\alpha) * \text{DIoU}
\end{equation}

\noindent where $alpha$ is a weight to balance the individual losses.


\section{Experimental setup}  \label{sec:experimental_setup}

This section details the experimentation conducted to assess the performance of the proposed methodology. First, we describe the two datasets considered, which are commonly used in object detection research. Next, we describe the evaluation process and the steps involved in the experimental procedure.

\subsection{Datasets}\label{AA}


We employed two well-known object detection datasets for the experimentation, which are briefly described as follows:

\begin{itemize}
    \item The Common Objects in Context (COCO) dataset~\cite{lin2014microsoft} is a large-scale object-focused dataset that has become a standard dataset for object detection and segmentation. It provides over 200,000 labeled images, covering 80 different object classes, each with its corresponding set of annotations, including bounding boxes, segmentation masks, or keypoint mapping.
    \item The Pascal Visual Object Classes (PascalVOC) dataset~\cite{everingham2010pascal} is an extensively used dataset in the computer vision community. While smaller than COCO, PascalVOC offers a high-quality dataset of around 10,000 images and 20 classes. There are different versions of this dataset, with PascalVOC 2007 and PascalVOC 2012 being the most commonly used among them. Each object in the dataset comes with a bounding box and a class label, along with additional metadata that are not required in this study. 
\end{itemize}


COCO was selected to pre-train the feature extractor, mainly because its diversity of objects and perspectives makes it a very comprehensive dataset. For object detection, the PascalVOC datasets were chosen because they are recognized benchmarks in the field and have been extensively used in prior studies, offering a good context for evaluating the hypothesis introduced with this research work. In particular, PascalVOC 2012 was used for training the object detector, and PascalVOC 2007 for evaluating it. Note that only those images that contain one single object are considered in both procedures. 


To train the object detector, we organized the experiments into two main groups. First, we used the so-called TINY dataset, which only includes the images corresponding to the five most represented classes in the PascalVOC 2012 dataset (i.e., the classes with more than 200 images per class): \textit{bird}, \textit{cat}, \textit{dog}, \textit{person}, and \textit{airplane}. This dataset was created to focus on classes with sufficient data available, allowing the training dataset to be balanced for those experiments requiring more labeled images. Subsequently, the entire PascalVOC 2012 dataset, referred to as the FULL dataset, was used with the 20 classes originally included in it. The purpose here was to present a more challenging scenario for evaluation. For both datasets, we allocated 80\% of the data for training and 20\% for validation. For testing, we used the entire PascalVOC 2007 dataset as the FULL dataset and generated its corresponding TINY dataset by selecting the images within the five classes mentioned above, following the procedure proposed by the authors of the PascalVOC Challenge~\cite{everingham2010pascal}.

\subsection{Evaluation}

This section describes the performance metrics used to assess the effectiveness of our proposal. Notice that, in object detection, both classification and localization tasks must be considered in terms of performance.

For the classification task, we measured the quality of our proposal through several levels of top-N classification accuracy. This metric quantifies the standard accuracy by determining if the true class corresponds to any of the N most probable classes predicted by the model. In this research, we considered three levels of accuracy, with $\text{N}=\{1,3,5\}$.

For object localization, we evaluated our methodology using the intersection over union~\cite{selvaraju2017grad}. This metric, which is considered a standard in the field, measures the extent of overlap between the ground truth bounding box ($B_{gt}$) and the predicted bounding box ($B$). It is defined as follows: 

\begin{equation}
\text{IoU} = \frac{|B \cap B_{gt}|}{|B \cup  B_{gt}|}.
\end{equation}


Based on this metric, we computed the mean IoU and the localization accuracy at a specified IoU threshold, $th$. Note that this localization accuracy is defined as the ability of the model to detect objects with an overlap greater than the threshold. To consider different degrees of overlap accuracy, we selected two thresholds, $th=\{0.5,0.7\}$.





\subsection{Experiment description}

The experiments were designed to simulate varying levels of labeled data availability, aiming to assess how effectively the proposed feature extractor, trained with SimCLR, performs in scenarios with limited labeled data compared to conventional pre-trained backbones. To achieve this, we used training sub-sets of different sizes, denoted as $n$. For the TINY dataset, which includes the five most frequent classes in PascalVOC, 
experiments were conducted using $n=\{10, 20, 50, 100, 200, 500\}$ labeled images per class. For the FULL dataset, which has a broader range of classes, the minimum number of images per class was set at 3, while the maximum was set at 200 to maintain dataset balance. Accordingly, experiments were carried out using $n=\{3, 5, 10, 20, 50, 100, 200\}$ labeled images per class.

All the images were resized to a spatial resolution of $224 \times 224$ pixels and normalized using a mean and standard deviation of 0.5. Regarding the training procedure, we used the following configuration for each phase:
\begin{itemize}
    \item \textit{SSL feature extraction.} The model was trained during a maximum of 200 epochs using the Adam~\cite{kingma2015adam} optimizer, with 0.0005 as the learning rate and 32 as the batch size. As suggested in~\cite{chen2020simple}, the data augmentation techniques applied were crop and resize, horizontal flip, color distortion, grayscale, Gaussian blur, and random erasing. 
    \item \textit{Object detection.} Adam was selected as the optimizer, with a learning rate of 0.001 and a batch size of 32. The model was trained during a maximum of 100 epochs using a value of $\alpha = 0.5$ in the loss function (see \ref{eq:object_loss}). The data augmentation techniques applied at this stage were random grayscale, Gaussian blur, and random erasing. 
\end{itemize}





\section{Results} \label{sec:results}

This section provides an in-depth analysis of the results obtained within this experimental scenario. In particular, we compared the performance of the proposed feature extractor trained using an SSL approach and referred to as \textit{SSL backbone}, against the backbone of an EfficientNetB1 trained on ImageNet and referred to as the \textit{Baseline}.

Table~\ref{tab:easy-results} and Table~\ref{tab:full-results} show the results obtained across all the experiments for the TINY and FULL datasets, respectively. These results are analyzed in the following sections.


\begin{table}[!htbp]
    \centering
    \caption{Comparative performance of the \textit{SSL backbone} against the \textit{Baseline} on the TINY test dataset.}
    \label{tab:easy-results}
    \resizebox{\linewidth}{!}{%
    \begin{tabular}{cc|cc|ccc}
        \toprule
        \multirow{2}{*}{$n$} & \multirow{2}{*}{Method} & \multicolumn{2}{c|}{Classification} & \multicolumn{3}{c}{Localization} \\ 
        & & Top-1 Acc & Top-3 Acc & Mean IoU & Acc IoU 0.5 & Acc IoU 0.7 \\  \midrule
        \multirow{2}{*}{10} & Baseline & 0.8262 & 0.9717 & 0.2594 & 0.1916 & 0.0440 \\
        & SSL & 0.4440 & 0.8094 & 0.4759 & 0.5215 & 0.2408 \\
        \multirow{2}{*}{20} & Baseline & 0.8806 & 0.9885 & 0.2818 & 0.2168 & 0.0408 \\
        & SSL & 0.4911 & 0.8293 & 0.4887 & 0.5455 & 0.2450 \\
        \multirow{2}{*}{50} & Baseline & 0.9277 & 0.9969 & 0.4095 & 0.3937 & 0.1099 \\
        & SSL & 0.5911 & 0.9288 & 0.5004 & 0.5602 & 0.2492 \\
        \multirow{2}{*}{100} & Baseline & 0.9361 & 0.9969 & 0.4778 & 0.5162 & 0.1539 \\
        & SSL & 0.6346 & 0.9403 & 0.5202 & 0.5969 & 0.2764 \\
        \multirow{2}{*}{200} & Baseline & 0.9518 & 0.9969 & 0.4895 & 0.5361 & 0.2000 \\
        & SSL & 0.6754 & 0.9550 & 0.5259 & 0.6115 & 0.2974 \\
        \multirow{2}{*}{500} & Baseline & 0.9560 & 0.9958 & 0.5261 & 0.5958 & 0.2963 \\ 
        & SSL & 0.7037 & 0.9675 & 0.5371 & 0.6157 & 0.3120 \\ \bottomrule
    \end{tabular}%
    }
\end{table}


\begin{table}[!htbp]
    \centering
    \caption{Comparative performance of the \textit{SSL backbone} against the \textit{Baseline} on the FULL test dataset.}
    \label{tab:full-results}
    \resizebox{\linewidth}{!}{%
    \begin{tabular}{cc|ccc|ccc}
        \toprule
        \multirow{2}{*}{$n$}& \multirow{2}{*}{Method} & \multicolumn{3}{c|}{Classification} & \multicolumn{3}{c}{Localization} \\ 
        &  & Top-1 Acc & Top-3 Acc & Top-5 Acc & Mean IoU & Acc IoU 0.5 & Acc IoU 0.7 \\ \midrule
        \multirow{2}{*}{3} & Baseline & 0.6259 & 0.8236 & 0.8880 & 0.1685 & 0.1206 & 0.0541 \\
        & SSL & 0.2223 & 0.4691 & 0.6215 & 0.4169 & 0.4080 & 0.1464 \\
        \multirow{2}{*}{5} & Baseline & 0.6248 & 0.8203 & 0.8984 & 0.1699 & 0.0983 & 0.0131 \\
        & SSL & 0.2615 & 0.4981 & 0.6412 & 0.4410 & 0.4659 & 0.1830 \\
        \multirow{2}{*}{10} & Baseline & 0.7160 & 0.8957 & 0.9483 & 0.2669 & 0.2108 & 0.0552 \\
        & SSL & 0.3124 & 0.6051 & 0.7378 & 0.4712 & 0.4992 & 0.2391 \\
        \multirow{2}{*}{20} & Baseline & 0.8225 & 0.9437 & 0.9716 & 0.3526 & 0.2911 & 0.0683 \\
        & SSL & 0.3905 & 0.6696 & 0.7946 & 0.4850 & 0.5390 & 0.2441 \\
        \multirow{2}{*}{50} & Baseline & 0.8547 & 0.9596 & 0.9776 & 0.4595 & 0.4861 & 0.1770 \\
        & SSL & 0.4702 & 0.7302 & 0.8454 & 0.4985 & 0.5516 & 0.2769 \\
        \multirow{2}{*}{100} & Baseline & 0.8755 & 0.9645 & 0.9803 & 0.4907 & 0.5483 & 0.2474 \\
        & SSL & 0.5210 & 0.7766 & 0.8684 & 0.5056 & 0.5576 & 0.2927 \\
        \multirow{2}{*}{200} & Baseline & 0.8815 & 0.9656 & 0.9820 & 0.5073 & 0.5680 & 0.2807 \\ 
        & SSL & 0.5904 & 0.8334 & 0.9126 & 0.5146 & 0.5685 & 0.3037 \\
        \bottomrule
    \end{tabular}%
    }
\end{table}

\subsection{Evaluation of the classifier performance}

The performance of the classifier within the object detectors was assessed using the Top-1, Top-3, and Top-5 accuracies. Note that the Top-5 was not taken into account in the experiments with the TINY dataset because it has only five classes and this metric would not give any insight into performance.

As can be seen in Tables~\ref{tab:easy-results} and~\ref{tab:full-results}, the \textit{Baseline} outperforms the \textit{SSL backbone} across all metrics and datasets. However, it is important to note that the \textit{Baseline} benefits from being pre-trained on ImageNet, a much larger dataset compared to the COCO dataset used to pre-train the \textit{SSL backbone}. In addition, the backbone of the \textit{Baseline} model was originally designed for classification tasks specifically but, regardless, the \textit{SSL} backbone still demonstrates good performance. These results show the great potential for significant improvements by pre-training the \textit{SSL backbone} on a larger unlabeled dataset and thus improving the downstream task architecture.

\begin{figure}[!htbp]
  \begin{subfigure}{0.5\linewidth}
    \includegraphics[width=\linewidth]{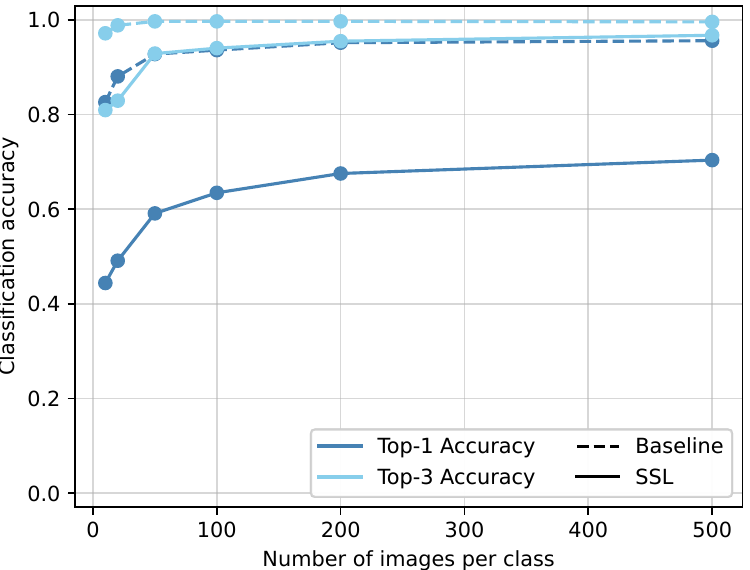}
    \caption{TINY}
  \end{subfigure}%
  \begin{subfigure}{0.5\linewidth}
    \includegraphics[width=\linewidth]{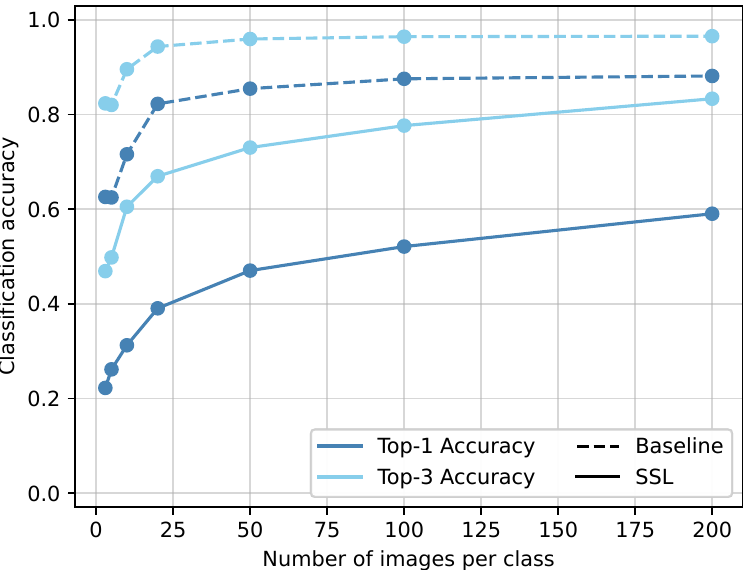}
    \caption{FULL}
  \end{subfigure}
  \caption{Top-1 and Top-3 classification accuracies for both datasets.}
  \label{fig:class-acc-top1-3}
\end{figure}

Despite not outperforming the \textit{Baseline} in the classification task, it is worth noting that, as can be seen in Fig.~\ref{fig:class-acc-top1-3}, the differences in Top-3 between the two methods are not so pronounced. In this figure, the evolution of the \mbox{Top-1} and \mbox{Top-3} metrics for both datasets can be seen.



\subsection{Evaluation of the object detector performance}

The performance of the object detection components was evaluated using the mean IoU and the localization accuracy at two IoU thresholds, 0.5 and 0.7. These three metrics provide a good understanding of the model's ability to accurately localize the objects within the images.

Analyzing the results depicted in Tables~\ref{tab:easy-results} and~\ref{tab:full-results}, we can see that the \textit{SSL backbone} outperforms the \textit{Baseline} model across all metrics and datasets. Our approach exhibits strong performance for both datasets, even when using a small number of images per class. This performance is especially remarkable given the architectural simplicity of the downstream detector, which consists of only a single layer following the feature extractor. This suggests that the feature extractor in the \textit{SSL backbone} is highly effective for object localization tasks, maintaining its performance even for minimal labeled data sizes, reinforcing its reliability and robustness. 

\begin{figure}[!htbp]
  \begin{subfigure}{0.5\linewidth}
    \includegraphics[width=\linewidth]{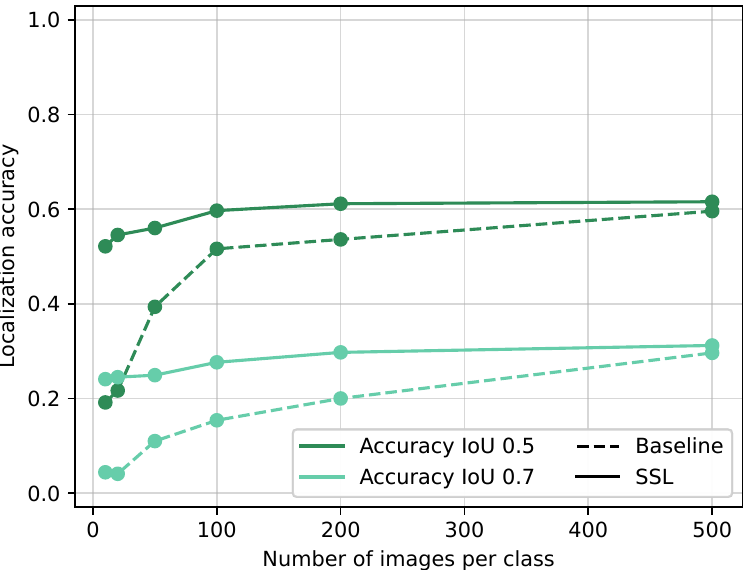}
    \caption{TINY}
  \end{subfigure}%
  \begin{subfigure}{0.5\linewidth}
    \includegraphics[width=\linewidth]{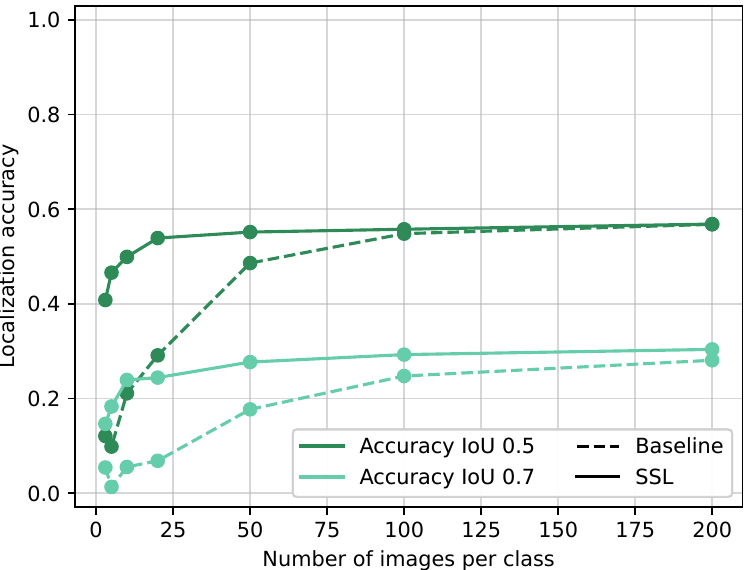}
    \caption{FULL}
  \end{subfigure}
  \caption{Localization accuracies at IoU 0.5 and 0.7 for both datasets.}
  \label{fig:loc-acc}
\end{figure}

Fig.~\ref{fig:loc-acc} illustrates the evolution of the localization accuracy at the two IoU thresholds considered.
As can be observed, the \textit{SSL backbone} exhibits good results even when the amount of labeled images is minimal. In fact, the difference with respect to the \textit{Baseline} increases as the amount of label data decreases. Therefore, these results demonstrate that the proposed approach is suited for applications with few labeled examples and can perform well in localization tasks. 

\subsection{Trade-off comparison}

In object detection, achieving a balanced performance between classification and localization presents a challenging task. For this reason, we examined the trade-offs between these two components across the two models considered. Our findings reveal that although the \textit{SSL backbone} does not perform as well as the \textit{Baseline} on classification metrics, it demonstrates superior performance in localization tasks.

The reason behind the \textit{SSL backbone} falling short when compared to the \textit{Baseline} in classification performance could be attributed to the difference in size between the datasets used to pre-train the feature extractor, as the ImageNet dataset used for the \textit{Baseline} has more than 14 million images while the COCO dataset used for the \textit{SSL backbone} has 40 times fewer examples. However, it is important to note that the classification performance of the \textit{SSL backbone} is still within acceptable ranges, particularly considering the simplicity of the architecture. On the other hand, the \textit{SSL backbone} significantly outperforms the \textit{Baseline} in localization tasks across all experiments,  with this advantage outweighing the deficiency in the classification task. 

\begin{figure*}[!htbp]
    \begin{subfigure}{\linewidth}
    \includegraphics[width=\linewidth]{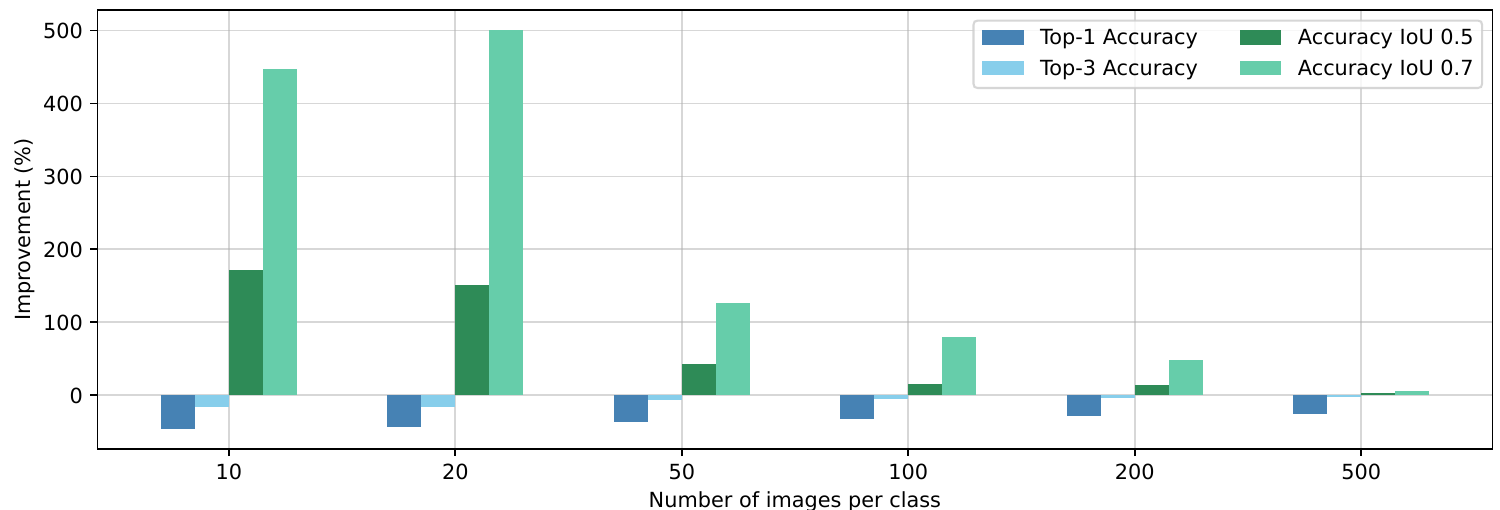}
    \caption{TINY}
  \end{subfigure}%
  
    \begin{subfigure}{\linewidth}
    \includegraphics[width=\linewidth]{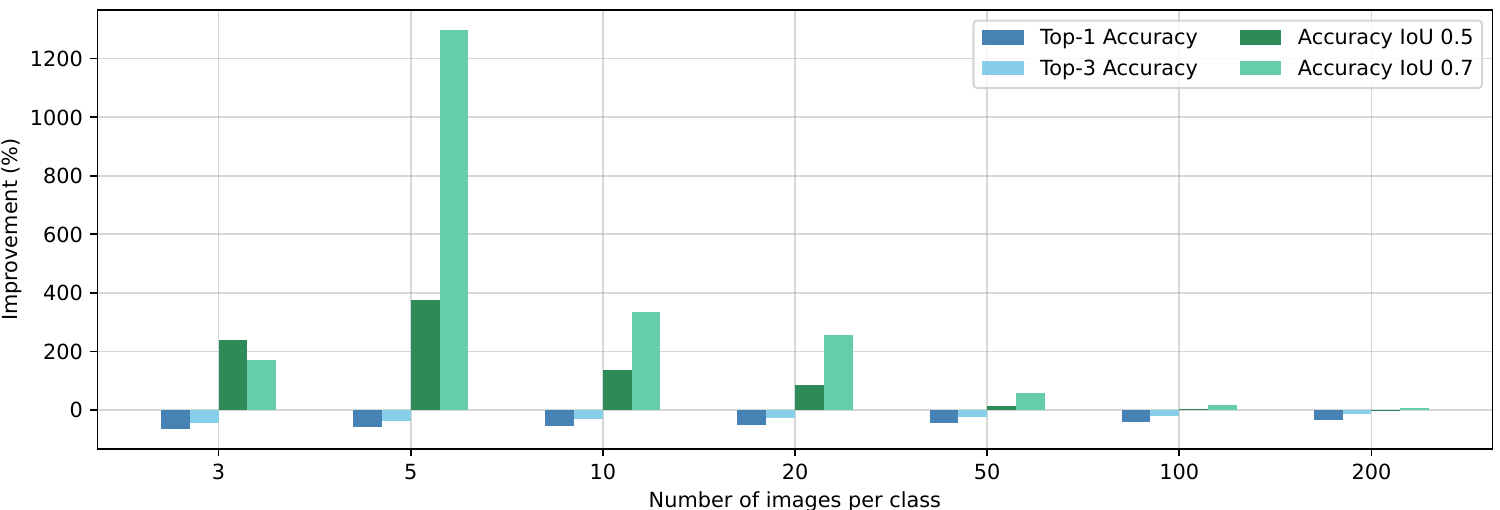}
    \caption{FULL}
  \end{subfigure}%
    \caption{Differences between the \textit{Baseline} and the \textit{SSL backbone} over the number of images per class for both datasets.}
    \label{fig:comparison-full}
\end{figure*}

Fig.~\ref{fig:comparison-full} shows the differences between the two approaches for four of the performance measures. Note that positive values correspond to an improvement of our proposed \textit{SSL backbone} over the \textit{Baseline}, while negative values mean a reduction in performance. As can be observed in both figures, despite our proposal being inferior in classification tasks, the gains in localization performance for the \textit{SSL backbone} are much higher than the losses. This makes our approach an effective alternative for object detection tasks, where both identification and localization capabilities are required.

\subsection{Visualizing feature extractor activations} 

To better understand the efficiency of our approach compared to the \textit{Baseline}, we use the Grad-CAM algorithm~\cite{selvaraju2017grad} to visualize the regions that the model pays attention to during the task. This algorithm works by backpropagating the gradient of the target label score to the feature map of the last convolutional layer. These gradients are used to generate a heatmap that highlights the most relevant regions of the input image that contributed to the final prediction. 

\begin{figure*}[!htbp]
  \centering
  \includegraphics[width=0.96\textwidth]{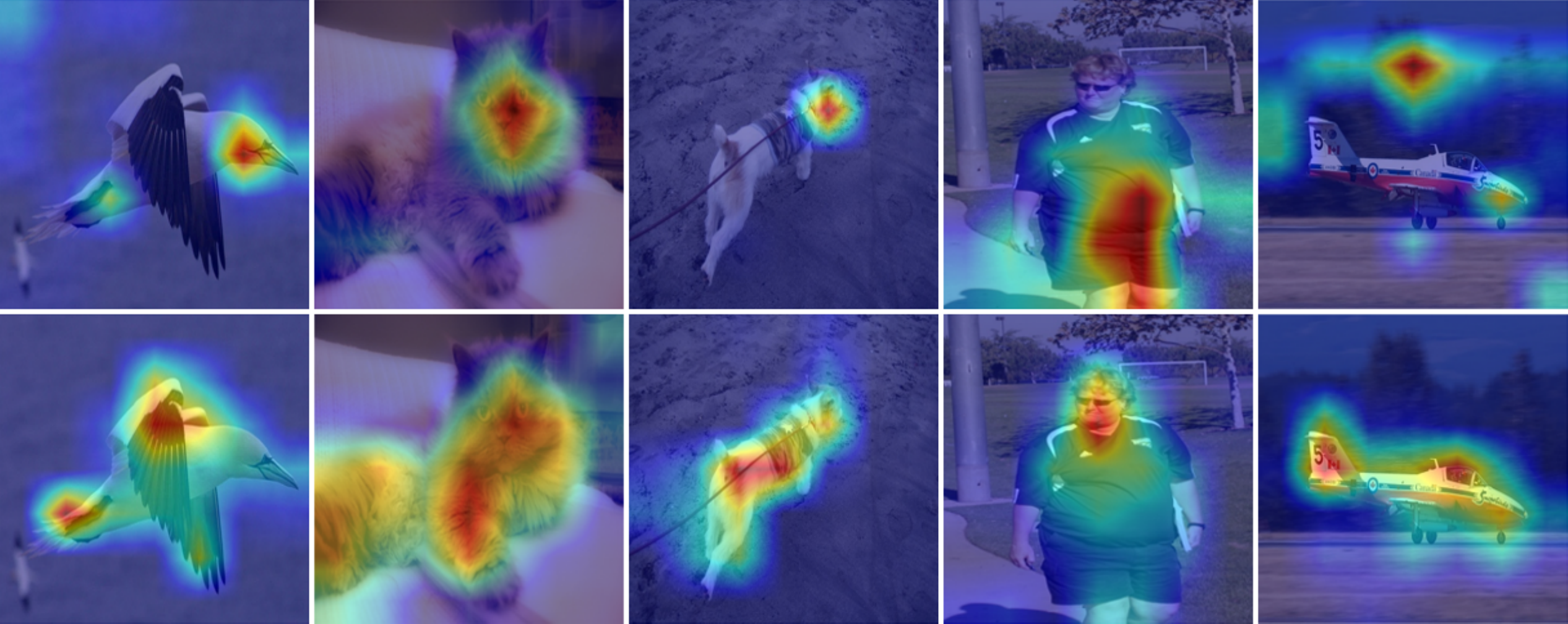}
  \caption{Heat maps generated by Grad-CAM for five representative samples from the PascalVOC datasets. The top row shows the activations from the \textit{Baseline} model, highlighting its focus on specific, fragmented regions of the object. The bottom row illustrates the activations of the \textit{SSL backbone}, capturing the entire object shape and its defining characteristics.}
  \label{fig:gradcam}
\end{figure*}

Fig.~\ref{fig:gradcam} reveals the different parts of the image that activate each model. It can be seen that the \textit{Baseline} model tends to focus on fragmented or very specific parts of the object, whereas the activations of the \textit{SSL backbone} cover the entire shape of the object more precisely, leading to a better understanding of the object's spatial context. By focusing on the totality of the object's shape, we can deduce that our approach is more suitable for tasks that require understanding the entirety of an object and not just the most salient characteristic. This also explains how the \textit{SSL backbone} showed significant improvements in localization metrics compared to the \textit{Baseline}, complementing our earlier findings and strengthening the hypothesis of using a \textit{SSL backbone} for object detection.

\section{Conclusions} \label{sec:conclusions}

The acquisition of images for real-world object detection applications can be straightforward, as it is possible to collect thousands of examples in a short time. However, labeling them becomes a long and expensive procedure. In this context, we proposed a self-supervised learning approach to train feature extractors that minimizes the required labeled data in deep learning applications and optimizes the training procedures. According to the experimentation carried out, the developed \textit{SSL backbone} performs well in object detection tasks, even with minimal amounts of labeled data. Therefore, our approach allows developers to use a major set of images to train a robust feature extractor and then adjust it with a smaller labeled set. This strategy can reduce the future need for big amounts of labeled data and complex architectures to achieve good performance, mitigating one important bottleneck for the industry nowadays.

Regarding the experimentation conducted to assess the effectiveness of this approach, we used different subsets of the PascalVOC dataset. Particularly, we examined the performance of the feature extractor using different quantities of labeled data, ranging from as few as 3 to as many as 500 labeled images per class. The results demonstrated that even with a very limited amount of labeled data, our \textit{SSL backbone} improves localization accuracy compared to the state-of-the-art feature extractors pre-trained on ImageNet. Furthermore, by using the heat maps generated through Grad-CAM, we can visualize that the features extracted with our \textit{SSL backbone} are considerably more relevant to the entire object, in contrast to those extracted by the \textit{Baseline} model. This suggests that our model not only learns to recognize objects but also refines its understanding of their spatial characteristics. 

While our main focus was on improving localization tasks, for real-world applications it becomes essential to improve the classification performance as well. To achieve this, future research includes pre-training the feature extractor with a larger unlabeled dataset, such as ImageNet. Additionally, it is important to note that the object detector architecture used in this work was intentionally simplified for experimentation purposes. Its objective was to highlight the capabilities of SSL pre-trained feature extractors, which is why we opted for a straightforward architecture. For this reason, we also plan to consider more complex architectures that would improve both classification and localization accuracies, while maintaining the advantages of training a high-performing model that requires fewer labeled data, thereby making the technology more accessible and efficient to both academia and industry. 

\bibliographystyle{IEEEtran}
\bibliography{IEEEabrv, biblio}

\end{document}